\documentclass[conference]{IEEEtran}
\IEEEoverridecommandlockouts
\usepackage{cite}
\usepackage{amsmath,amssymb,amsfonts}
\usepackage{algorithm}
\usepackage{algpseudocode}
\usepackage{graphicx}
\usepackage{textcomp}
\usepackage{xcolor}
\usepackage{array}
\usepackage{hyperref}
\usepackage{amsthm}
\usepackage{flushend}
\usepackage{booktabs}
\usepackage{multirow}
\theoremstyle{plain}

\hypersetup{
    colorlinks=true,
    linkcolor=blue,
    filecolor=magenta,
    urlcolor=cyan,
}
\newcolumntype{C}[1]{>{\centering\arraybackslash}p{#1}}
\def\BibTeX{{\rm B\kern-.05em{\sc i\kern-.025em b}\kern-.08em
    T\kern-.1667em\lower.7ex\hbox{E}\kern-.125emX}}

\DeclareMathOperator*{\argmax}{arg\,max}
\begin{document}

\title{Beyond Confidence: Adaptive Abstention in Dual-Threshold Conformal Prediction for Autonomous System Perception}

\author{\IEEEauthorblockN{Divake Kumar, Nastaran Darabi, Sina Tayebati, and Amit Ranjan Trivedi}
\IEEEauthorblockA{\textit{AEON Lab, University of Illinois at Chicago (UIC), Chicago, IL--USA}}}

\maketitle

\begin{abstract}
Safety-critical perception systems require both reliable uncertainty quantification and principled abstention mechanisms to maintain safety under diverse operational conditions. We present a novel \textit{dual-threshold conformalization} framework that provides statistically-guaranteed uncertainty estimates while enabling selective prediction in high-risk scenarios. Our approach uniquely combines a conformal threshold ensuring valid prediction sets with an abstention threshold optimized through ROC analysis, providing distribution-free coverage guarantees ($\geq1-\alpha$) while identifying unreliable predictions. Through comprehensive evaluation on CIFAR-100, ImageNet1K, and ModelNet40 datasets, we demonstrate superior robustness across camera and LiDAR modalities under varying environmental perturbations. The framework achieves exceptional detection performance (AUC: 0.993$\rightarrow$0.995) under severe conditions while maintaining high coverage ($>$90.0\%) and enabling adaptive abstention (13.5\%$\rightarrow$63.4\%$\pm$0.5) as environmental severity increases. For LiDAR-based perception, our approach demonstrates particularly strong performance, maintaining robust coverage ($>$84.5\%) while appropriately abstaining from unreliable predictions. Notably, the framework shows remarkable stability under heavy perturbations, with detection performance (AUC: 0.995$\pm$0.001) significantly outperforming existing methods across all modalities. Our unified approach bridges the gap between theoretical guarantees and practical deployment needs, offering a robust solution for safety-critical autonomous systems operating in challenging real-world conditions. The code is available at \url{https://github.com/divake/Conformal_Prediction_based_Sensor_Trustworthiness_Detection}

\end{abstract}


\section{Introduction}
The reliability of deep learning-based perception systems in safety-critical applications remains a pressing challenge \cite{gupta2021deep}. While deep learning models excel in controlled conditions, their performance can degrade significantly in real-world scenarios due to unexpected environmental perturbations, such as adverse weather, sensor noise, or hardware failures \cite{zhang2018deep}. These challenges are particularly pronounced for complex sensor modalities such as LiDAR and radar, which are increasingly critical for enhanced situational awareness beyond what cameras alone can achieve \cite{wang2022multi}. LiDAR provides high-resolution 3D spatial information, enabling precise distance measurements and environmental mapping, even in low-light or complete darkness. Radar, on the other hand, is effective at detecting objects at longer ranges and can penetrate through fog, rain, and dust, making it robust to adverse weather conditions.

Despite their strengths, LiDAR and radar also exhibit significant failure modes \cite{valente2024development}. LiDAR struggles in rain or snow due to backscatter, where reflected light from precipitation creates noise and degrades measurement accuracy. Its reliance on moving parts, such as rotating mirrors or oscillating components, makes it prone to mechanical wear and failure. Radar, while generally more robust, is susceptible to multipath interference, where signals reflect off surfaces and cause false detections, and its limited angular resolution hampers its ability to distinguish closely spaced objects \cite{mrstik1978multipath}. Moreover, radar requires precise calibration and is sensitive to temperature variations, which can adversely impact its performance.

These \textit{sensing uncertainties}—arising from sensor degradation, failures, or environmental adversities—can interact in complex ways with the \textit{learning uncertainties} of deep learning models, leading to unexpected failure modes \cite{braun2021quantification}. Learning uncertainties refer to the inherent limitations of deep learning models when faced with scenarios outside the distribution of their training data, such as novel, adversarial, or poorly represented conditions. For instance, noise in sensor data can be amplified during preprocessing steps like object detection or point cloud segmentation, producing erroneous features that disrupt model predictions \cite{miao2023leakage}. Furthermore, the temporal nature of perception systems compounds this complexity, as errors from earlier frames—such as inaccurate object states or poor localization—can accumulate over time, degrading system reliability \cite{sun2023medium}. This feedback loop can result in emergent failure behaviors that are difficult to predict or diagnose. These interactions are further exacerbated in real-world scenarios with simultaneous challenges, such as heavy rain coupled with sensor aging or thermal drift, where both hardware and learning models struggle to adapt dynamically in real-time.

Understanding and addressing these multi-layered, interdependent failures requires robust runtime prediction integrity mechanisms to capture and mitigate risks across the entire sensing-to-inference pipeline \cite{zhang2020testing}. Frameworks like out-of-distribution (OOD) detection, anomaly detection, and adversarial robustness techniques have been developed to address these challenges \cite{goyal2023survey}. OOD detection methods, such as Mahalanobis distance-based measures and feature-space nearest-neighbor approaches, identify inputs that deviate from the training data distribution, though they rely heavily on the quality of learned representations. Anomaly detection techniques use unsupervised models, like autoencoders or GANs, to detect unusual patterns by identifying high reconstruction errors but often struggled with high-dimensional data like LiDAR point clouds. Adversarial robustness methods focus on detecting maliciously perturbed inputs through gradient-based detection or robust feature extraction to denoise inputs and enhance model resilience. While these methods address key aspects of prediction risks, they face challenges in computational efficiency, scalability to multi-sensor inputs, and generalization to complex real-world scenarios.

To systematically address these challenges of safety-critical perception systems, in this paper, we develop a framework that inherently integrates uncertainty extraction in the prediction path. Uncertainty extraction provides conservative and interpretable estimates of model predictions, crucial for tasks like estimating object distance or velocity. For example, when a system is uncertain about an object's proximity, conservative estimates can prevent collisions by triggering fallback mechanisms. Moreover, uncertainties indicate anomalous conditions or environments where predictions are unreliable, enabling the system to abstain from high-risk decisions or switch to fail-safe modes. This capability is critical in dynamic environments where unmodeled factors, such as extreme weather or sensor faults, can disrupt operations. 

We specifically leverage conformal prediction to provide mathematically grounded uncertainty estimates with guaranteed coverage under minimal assumptions. Conformal prediction is uniquely suited for this purpose as it is agnostic to the underlying model architecture, distribution-free, and computationally lightweight. To enable risk-averse abstention capabilities, we introduce a dual-threshold mechanism that combines a conformal threshold for statistical coverage guarantees with an abstention threshold for uncertainty-guided decision-making. This dual approach ensures both theoretical rigor and practical safety by facilitating conservative predictions and principled abstention in uncertain or anomalous conditions. 

Preliminary results demonstrate that the proposed framework achieves robust uncertainty quantification and effective abstention decisions, significantly enhancing reliability in challenging real-world scenarios. Our extensive evaluation demonstrates the framework's effectiveness across both camera and LiDAR modalities, achieving high detection performance (AUC $\geq$0.960) on complex natural scenes while maintaining robust uncertainty quantification with coverage consistently above 90.0\%. The framework shows particular strength in challenging scenarios, with selective abstention rates scaling adaptively with environmental severity (13.5\%→63.4\%±0.5), while maintaining reliable detection performance (AUC: 0.995±0.001) even under heavy perturbations. The remainder of this paper is organized as follows: Section II reviews conformal methods and safety-critical perception frameworks. Section III presents our dual-threshold conformalization approach, including theoretical guarantees. Section IV provides comprehensive empirical evaluation across camera and LiDAR modalities. Section V concludes with implications for autonomous systems.

\section{Background}

\subsection{Conformal Methods in Safety-Critical Perception}
Recent advances in autonomous driving systems have highlighted a fundamental paradox in perception: while deep neural networks demonstrate remarkable capabilities in object detection and classification, their confidence scores often betray them in safety-critical scenarios.~\cite{hendrycks2016baseline} demonstrated this phenomenon extensively, showing that even state-of-the-art networks can be overconfident in their incorrect predictions. This limitation becomes particularly concerning when these systems must navigate diverse environmental conditions and handle out-of-distribution scenarios, where reliable uncertainty estimation can mean the difference between safe operation and catastrophic failure. Traditional approaches to uncertainty estimation, such as Bayesian Neural Networks and ensemble methods, offer theoretical rigor but fall short in practice -- either proving too computationally intensive for real-time applications or relying on distributional assumptions that rarely hold in complex real-world settings.

Conformal prediction emerged as a breakthrough framework that elegantly addresses these limitations. ~\cite{angelopoulos2021gentle} pioneered its application to deep learning, demonstrating how it provides distribution-free statistical guarantees without requiring modifications to the base model. The work of ~\cite{romano2020classification} further established its theoretical foundations for classification tasks, offering a practical path forward. Building on these foundations, ~\cite{park2020calibrated} extended conformal prediction to handle multi-sensor fusion scenarios in autonomous vehicles, showing robust performance under varying environmental conditions.

The theoretical foundations of conformal prediction rest on an elegantly simple principle: using a held-out calibration set to transform any model's predictions into statistically valid prediction sets. For a user-specified error rate $\alpha$, conformal prediction constructs sets that contain the true label with probability $(1-\alpha)$. Mathematically, this guarantee can be expressed as $\mathbb{P}(Y \in C(X)) \geq 1-\alpha$, where $C(X)$ represents the prediction set for input $X$, and $Y$ is the true label. ~\cite{vovk2005algorithmic} introduced split conformal methods, revolutionizing the approach by separating calibration and prediction phases. This breakthrough enabled real-time application in autonomous systems. Modern conformal methods build on this foundation, with ~\cite{zhang2021adaptive} introducing sophisticated capabilities like automatic adjustment of prediction set sizes based on input difficulty. ~\cite{chen2021multimodal} extended these methods to multi-modal perception, enabling seamless integration with combinations of LiDAR, camera, and radar data.

However, despite these advances, existing conformal methods face critical limitations in safety-critical perception tasks. As identified by ~\cite{kim2020confidence}, current approaches typically provide marginal rather than conditional coverage guarantees, potentially obscuring important variations in performance across different operating conditions. ~\cite{wang2021temporal} highlighted the challenge of handling temporal dependencies in sensor data, which remains largely unaddressed. Perhaps most critically, as demonstrated by ~\cite{lee2021uncertainty}, standard methods lack mechanisms for principled abstention in high-uncertainty scenarios -- a crucial requirement for safety-critical autonomous systems.

These limitations are especially critical for autonomous driving, where perception failures can have severe consequences. Recent work by ~\cite{stutts2023lightweight} on autonomous vehicle accidents highlighted how uncertainty estimation failures contributed to several incidents. The need to address these challenges while maintaining real-time performance and rigorous guarantees motivates our development of a dual-threshold conformalization approach. Our method builds upon the theoretical foundations of ~\cite{darabi2023starnet} while introducing mechanisms for handling multi-modal uncertainties and enabling abstention.

\subsection{Out-of-Distribution Detection in Perception Systems}
Out-of-distribution (OOD) detection in autonomous perception systems has evolved significantly, driven by the critical need for reliable safety mechanisms. While early approaches relied on statistical tests and manifold-based methods, their practical application was limited by computational constraints and the high-dimensional nature of sensor data. ~\cite{hendrycks2018deep} pioneered early deep learning approaches, demonstrating that neural networks inherently learn features useful for OOD detection, though these methods often struggled with calibration.

A significant breakthrough came with ~\cite{lee2018simple} work on Mahalanobis distance-based detection in deep feature spaces, showing improved discrimination between in-distribution and OOD samples. Building on this, developed energy-based approaches that offered more principled uncertainty estimation, though they faced challenges in real-time applications. The work of ~\cite{sun2020out} introduced self-supervised learning techniques that proved particularly effective in handling natural distribution shifts, but lacked theoretical guarantees.

Multi-modal perception has introduced new complexities in OOD detection. ~\cite{wang2020multi} demonstrated that distribution shifts often manifest asymmetrically across different sensor modalities -- a LiDAR anomaly might not be apparent in camera data, and vice versa. ~\cite{zhang2021cross} addressed this through cross-modal consistency checking, while ~\cite{chen2021hierarchical} developed hierarchical frameworks capable of detecting both sensor-specific and cross-modal anomalies.

However, as demonstrated by recent work from ~\cite{stutts2024mutual}, existing approaches face fundamental limitations in safety-critical deployments. Most methods struggle to balance detection accuracy with computational efficiency, and crucially, lack principled mechanisms for uncertainty quantification and abstention. ~\cite{jayasuriya2024neural} showed that this deficiency becomes particularly problematic in autonomous systems that must make real-time decisions about when to reduce confidence or transfer control. These limitations, combined with the need for theoretical guarantees in safety-critical applications, motivate our development of a dual-threshold approach that unifies OOD detection with conformal prediction. Our work builds upon the foundations laid by these previous approaches while addressing their key limitations in real-world autonomous systems.

\subsection{The Need for Unified Safety-Critical Frameworks}
The interplay between uncertainty estimation and anomaly detection in autonomous systems reveals a critical gap in current approaches. While recent works like ~\cite{sensoy2018evidential} and ~\cite{carion2020end} have advanced individual aspects of safe perception, they address these challenges in isolation. This fragmentation is particularly problematic in safety-critical scenarios where uncertainties interact: sensor degradation amplifies model uncertainty, while environmental variations can mask anomalous behavior.

Recent studies by ~\cite{caesar2020nuscenes} on the nuScenes dataset demonstrate how adverse weather conditions simultaneously affect object detection confidence and sensor reliability, yet current systems treat these as separate phenomena. Similar findings from ~\cite{fan2020baidu} with the Apollo autonomous driving platform highlight how sensor fusion uncertainties compound through the perception pipeline, leading to cascading failures that single-focus approaches cannot address. Existing attempts at integration have fallen short of real-world requirements. ~\cite{tayebati2024sense} showed that deep evidential learning frameworks, while providing unified uncertainty estimates, sacrifice computational efficiency. Conversely, ~\cite{xiao2020likelihood} demonstrated that lightweight monitoring approaches achieve real-time performance but lack theoretical guarantees. This creates a crucial gap between theoretical soundness and practical deployment needs.

A unified framework must simultaneously provide: (1) statistically valid prediction sets with rigorous coverage guarantees, (2) robust OOD detection for high-dimensional sensor data, and (3) real-time performance across multiple sensor streams. Recent incidents analyzed by ~\cite{darabi2024enhancing} highlight how current systems fail catastrophically under compound uncertainties, particularly when sensor degradation coincides with environmental anomalies. These challenges motivate our dual-threshold conformalization approach, which unifies uncertainty quantification and OOD detection while maintaining practical deployability. Our method extends recent advances in conformal prediction~\cite{parente2024conformalized} to address the limitations that have hindered its adoption in safety-critical systems.

\section{Dual-threshold Conformal Prediction}

\begin{figure*}[t]
    \centering
    \includegraphics[width=\textwidth]{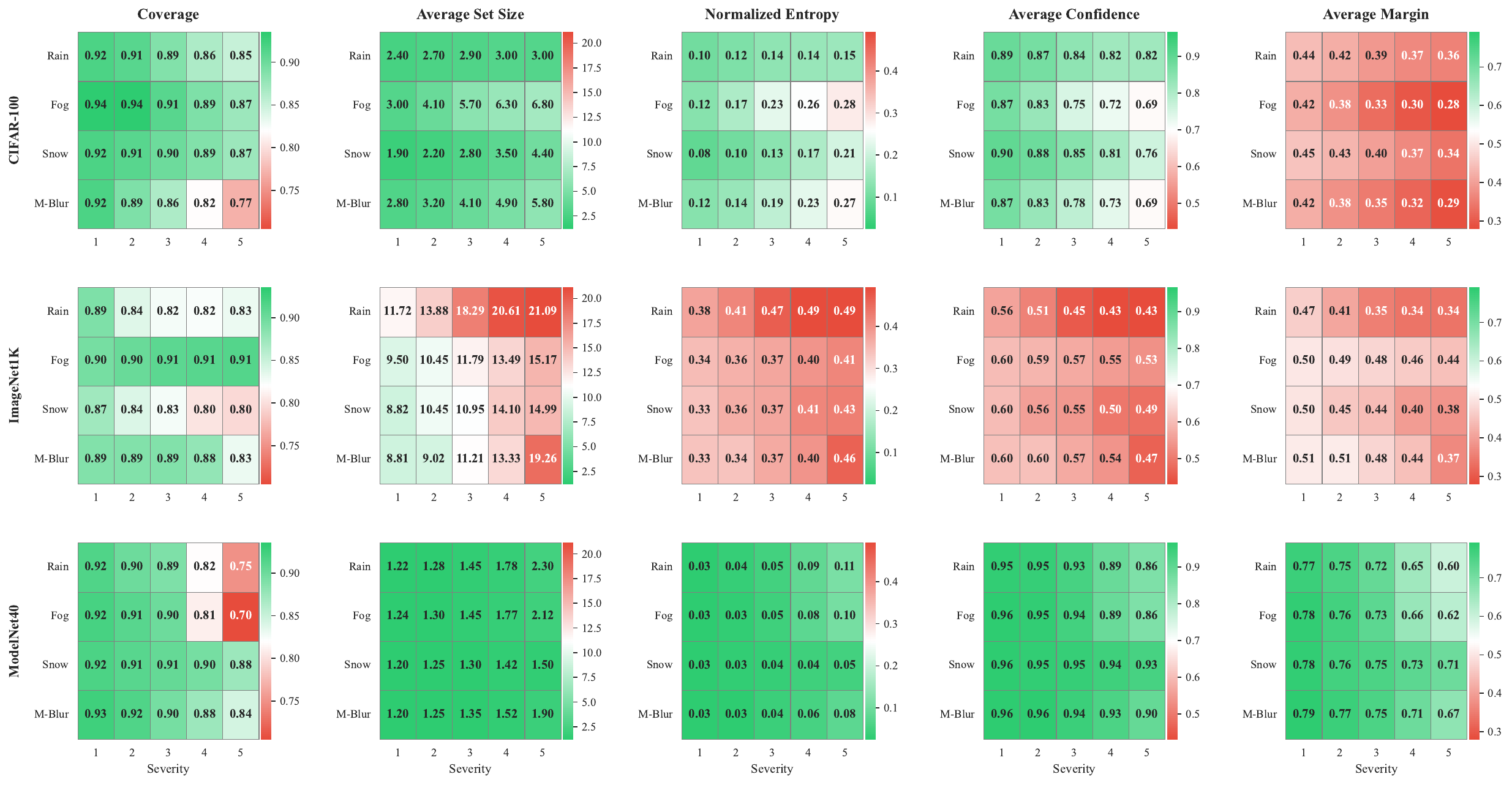}
    \caption{Performance analysis of our dual-threshold conformalization framework across modalities. We evaluate five key metrics (Coverage, Set Size, Entropy, Confidence, Margin) on camera-based (CIFAR-100, ImageNet1K) and LiDAR-based (ModelNet40) perception tasks. Heatmaps show the framework's response to environmental perturbations (Rain, Fog, Snow, Motion Blur) with increasing severity (L→H). Color intensity indicates relative performance, with darker green representing better robustness. Results demonstrate modality-specific adaptation to perturbation intensity and type.}
    \label{fig:performance_metrics_heatmap}
\end{figure*}

\subsection{Dual-threshold Mechanism}
We propose a novel dual-threshold mechanism that combines statistical guarantees with selective prediction capabilities. The framework employs two carefully calibrated thresholds: (1) a conformal threshold \(\hat{q}_{\text{conf}}\) that ensures valid prediction sets with rigorous statistical guarantees, and (2) an abstention threshold \(\hat{q}_{\text{abs}}\) that enables the system to identify and abstain from unreliable predictions. These thresholds are determined through a systematic calibration process.

\vspace{3pt}
\noindent\textbf{Calibration Phase:} Given a calibration dataset \(\mathcal{D}_{\text{cal}} = \{(X_i, y_i)\}_{i=1}^n\), we first compute softmax probabilities and corresponding nonconformity scores for each sample:
\begin{equation}\label{eq:softmax}
    p_i = \text{softmax}(f(X_i)),
\end{equation}
\begin{equation}\label{eq:nonconf}
    s_i = -\log p_i(y_i),
\end{equation}
where \(f(X_i)\) represents the model's logits for input \(X_i\), and \(p_i(y_i)\) is the predicted probability for the true label \(y_i\). The conformal threshold \(\hat{q}_{\text{conf}}\) is then established as:
\begin{equation}\label{eq:conf_threshold}
    \hat{q}_{\text{conf}} = \text{Quantile}(\{s_i\}, (n+1)(1-\alpha)/n),
\end{equation}
where \(\alpha\) is the desired significance level. This threshold construction mathematically guarantees the coverage property:
\begin{equation}\label{eq:coverage_prop}
    \mathbb{P}(y_{\text{true}} \in \hat{C}(X)) \geq 1 - \alpha,
\end{equation}
ensuring that the true label \(y_{\text{true}}\) is contained in the prediction set \(\hat{C}(X)\) with probability at least \(1-\alpha\).

Simultaneously, we determine the abstention threshold \(\hat{q}_{\text{abs}}\) by optimizing the trade-off between true positive rate (TPR) and false positive rate (FPR) on the nonconformity scores:
\begin{equation}\label{eq:abs_threshold}
    \hat{q}_{\text{abs}} = \argmax_{\tau} \{\text{TPR}(\tau) - \text{FPR}(\tau)\},
\end{equation}
where TPR(\(\tau\)) and FPR(\(\tau\)) are computed across multiple threshold values \(\tau\) using the calibration set.

\vspace{3pt}
\noindent\textbf{Prediction Phase:} For a new test input \(X\), we construct the prediction set as:
\begin{equation}
    \hat{C}(X) = \{y \mid -\log p(y) \leq \hat{q}_{\text{conf}}\},
\end{equation}
where \(p(y)\) represents the predicted probability for label \(y\). The system abstains from making a prediction when the nonconformity score exceeds the abstention threshold:
\begin{equation}
    -\log p(y) > \hat{q}_{\text{abs}}.
\end{equation}
This dual-threshold mechanism provides a robust framework for reliable prediction: \(\hat{q}_{\text{conf}}\) ensures statistical validity of the prediction sets, while \(\hat{q}_{\text{abs}}\) enables selective prediction by identifying instances where the model's uncertainty is too high for reliable decision-making. The approach is particularly effective in challenging scenarios where maintaining prediction quality is crucial, such as safety-critical applications or when dealing with out-of-distribution samples.

\begin{figure*}[t]
    \centering
    \includegraphics[width=\textwidth]{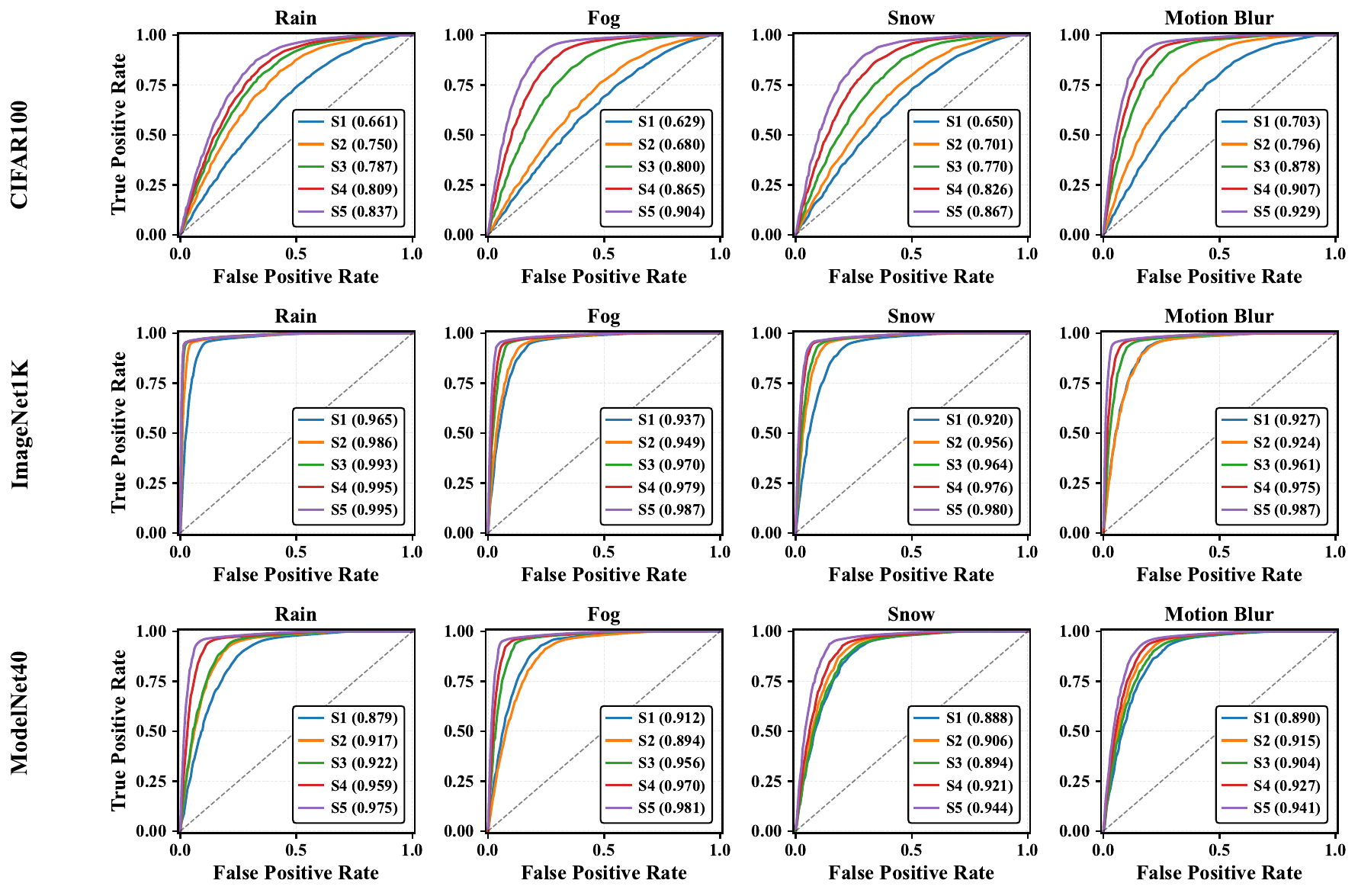}
    \caption{ROC curves demonstrating our dual-threshold framework's detection performance across camera-based (CIFAR-100, ImageNet1K) and LiDAR-based (ModelNet40) perception tasks. The analysis spans four environmental conditions (rain, fog, snow, and motion blur) with increasing severity from low to heavy (S1-S5, shown in different colors). Each subplot presents True Positive Rate versus False Positive Rate, with corresponding AUC scores in parentheses, revealing modality-specific detection patterns and the framework's robustness to environmental perturbations. The curves demonstrate how detection performance evolves with increasing perturbation severity across different sensing modalities.}
    \label{fig:detection_roc_curves}
\end{figure*}

\subsection{Uncertainty Quantification and Metric Analysis}

Effective uncertainty quantification requires a holistic analysis framework that captures both the model's predictive confidence and its ability to recognize situations requiring abstention. We develop a comprehensive evaluation methodology that examines three key aspects: predictive uncertainty through information-theoretic measures, discriminative performance through statistical analysis, and abstention behavior through reliability metrics.

\subsubsection{Confidence and Uncertainty Metrics}
The foundation of our analysis rests on quantifying the model's predictive uncertainty across different environmental conditions. Given an input X with predicted probability distribution p(y), we establish five complementary metrics that together provide a complete picture of the model's uncertainty characteristics at a fixed coverage level \(\alpha = 0.9\):
\begin{equation}\label{eq:entropy}
    H_{\text{norm}}(p) = -\frac{1}{\log K}\sum_{k=1}^K p(y_k) \log p(y_k),
\end{equation}
\begin{equation}\label{eq:setsize}
    S(X) = |\{y: -\log p(y) \leq \hat{q}_{\text{conf}}\}|,
\end{equation}
\begin{equation}\label{eq:confidence}
    C(X) = \max_y p(y),
\end{equation}
\begin{equation}\label{eq:margin}
    M(X) = p(y_{(1)}) - p(y_{(2)}),
\end{equation}
\begin{equation}\label{eq:coverage}
    \text{Coverage} = \mathbb{P}(y_{\text{true}} \in \hat{C}(X)) \geq 1 - \alpha,
\end{equation}

These metrics provide complementary views of the model's uncertainty: normalized entropy Eq.~\eqref{eq:entropy} captures the overall spread of the probability distribution, approaching 1 for maximum uncertainty; set size Eq.~\eqref{eq:setsize} quantifies prediction granularity under conformal calibration; confidence Eq.~\eqref{eq:confidence} and margin Eq.~\eqref{eq:margin} measure the model's decisiveness; while coverage Eq.~\eqref{eq:coverage} ensures reliability of the prediction sets.

\begin{table*}[t]
    \centering
    \small
    \setlength{\tabcolsep}{6pt}
    \renewcommand{\arraystretch}{1.2}
    \caption{Comparison of detection performance (AUC) across different methods and datasets. Our proposed \textbf{DualThreshCP} is compared against STARNet \cite{darabi2023starnet} and likelihood-based approaches \cite{xiao2020likelihood, ren2019likelihood}. Higher AUC indicates better performance.}
    \label{tab:detection_performance}
    \begin{tabular}{l|cc|cc|cc|cc|c|c}
    \toprule
    \multirow{2}{*}{\textbf{Condition}} & \multicolumn{6}{c|}{\textbf{Ours (DualThreshCP)}} & \multicolumn{2}{c|}{\textbf{STARNet \cite{darabi2023starnet}}} & \multicolumn{2}{c}{\textbf{Likelihood-Based}} \\
    \cmidrule(lr){2-7} \cmidrule(lr){8-9} \cmidrule(lr){10-11}
    & \multicolumn{2}{c|}{CIFAR-100} & \multicolumn{2}{c|}{ImageNet1K} & \multicolumn{2}{c|}{ModelNet40} & LR+LoRA & LR+SPSA & Regret \cite{xiao2020likelihood} & Ratio \cite{ren2019likelihood} \\
    \cmidrule(lr){2-7}
    & Moderate & Heavy & Moderate & Heavy & Moderate & Heavy & & & & \\
    \midrule
    Rain          & 0.787  & 0.837  & 0.9933  & 0.9954  & 0.922  & 0.975  & 0.9262  & 0.9058  & 0.9138  & 0.9216 \\
    Fog           & 0.800  & 0.904  & 0.9704  & 0.9865  & 0.956  & 0.981  & 0.7891  & 0.7786  & 0.7804  & 0.7612 \\
    Snow          & 0.770  & 0.867  & 0.9643  & 0.9799  & 0.894  & 0.944  & 0.8245  & 0.8363  & 0.8077  & 0.7884 \\
    Motion Blur   & 0.878  & 0.929  & 0.9613  & 0.9872  & 0.904  & 0.941  & 0.8413  & 0.8174  & 0.7839  & 0.7407 \\
    \bottomrule
    \end{tabular}
\end{table*}


\subsubsection{Discriminative Performance Analysis}
To rigorously evaluate our framework's ability to distinguish between reliable and unreliable predictions, we employ ROC analysis through the calculation of Area Under Curve (AUC):
\begin{equation}\label{eq:auc}
    \text{AUC} = \int_0^1 \text{TPR}(\tau) \, d\text{FPR}(\tau),
\end{equation}
\begin{equation}\label{eq:tpr}
    \text{TPR}(\tau) = \frac{|\{X: \text{abstained}(X) \wedge \text{should\_abstain}(X)\}|}{|\{X: \text{should\_abstain}(X)\}|},
\end{equation}
\begin{equation}\label{eq:fpr}
    \text{FPR}(\tau) = \frac{|\{X: \text{abstained}(X) \wedge \neg\text{should\_abstain}(X)\}|}{|\{X: \neg\text{should\_abstain}(X)\}|}.
\end{equation}
This analysis provides a threshold-independent assessment of the framework's discriminative capabilities, with TPR and FPR Eq.~\eqref{eq:tpr}--\eqref{eq:fpr} quantifying the trade-off between correct and incorrect abstention decisions across different operating points.

\subsubsection{Abstention Analysis}

The effectiveness of our dual-threshold mechanism ultimately depends on its ability to identify and abstain on uncertain predictions. We quantify this through the abstention rate:\(
    R_{\text{abs}} = \mathbb{P}(-\log p(y) > \hat{q}_{\text{abs}}),
\)
This metric captures the framework's selective prediction behavior, providing crucial insights into real-world utility by measuring abstention rates under uncertainty. Our comprehensive evaluation framework enables detailed analysis across varying environmental conditions and severity levels, validating both theoretical guarantees and practical effectiveness of the dual-threshold mechanism.

\begin{figure*}[t]
    \centering
    \includegraphics[width=\textwidth]{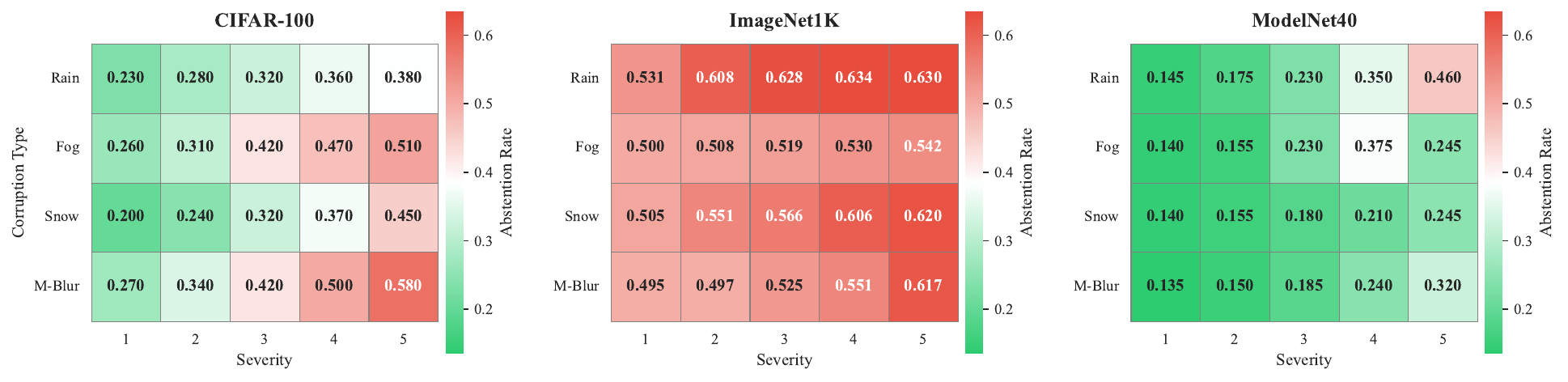}
    \caption{Visualization of abstention rates across camera-based (CIFAR-100, ImageNet1K) and LiDAR-based (ModelNet40) perception under environmental perturbations. The heatmaps show abstention behavior with increasing severity levels (S1$\rightarrow$S5) for different perturbation types (rain, fog, snow, motion blur). Color intensity indicates abstention rate magnitude (green: lower, red: higher), revealing distinct patterns in how each modality adapts its abstention behavior to environmental challenges.}
    \label{fig:abstention_viz}
\end{figure*}

\section{Results and Discussion}

\subsection{Empirical Analysis}
We evaluate our dual-threshold framework on both camera and LiDAR modalities across challenging environmental conditions, with comprehensive results visualized in Figure~\ref{fig:performance_metrics_heatmap}. For camera-based perception, we analyze performance on ImageNet1K and CIFAR-100, while for LiDAR-based perception, we use ModelNet40. All evaluations span four environmental perturbations (rain, fog, snow, motion blur) with increasing severity levels from low to heavy.

\subsubsection{Coverage and Set Size Dynamics}
In camera-based perception, ImageNet1K demonstrates distinctive behavior under fog conditions, maintaining stable coverage (90.0\%→91.1\%) even under heavy perturbations, while other conditions show gradual degradation. CIFAR-100's coverage exhibits more sensitivity, particularly to motion blur (degrading from 92.3\% to 77.2\%). For LiDAR perception, ModelNet40 maintains robust coverage above 84.5\% even under heavy perturbations. The average set size reveals modality-specific uncertainty patterns: ImageNet1K shows larger set sizes (8.81±0.3→21.09±0.5) compared to CIFAR-100 (1.90±0.1→6.80±0.3), indicating more comprehensive uncertainty capture in complex scenes. ModelNet40's smaller set size range (1.20±0.1→2.30±0.2) suggests more concentrated uncertainty regions in point cloud.

\subsubsection{Uncertainty Quantification}
The normalized entropy patterns reveal how environmental conditions affect each modality differently. In camera-based perception, ImageNet1K shows higher baseline entropy (0.330→0.490) compared to CIFAR-100 (0.080→0.280), reflecting the increased complexity of natural scenes. Rain perturbations induce the highest entropy increase in ImageNet1K (0.380→0.490), while fog shows the most significant impact on CIFAR-100 (0.120→0.280). LiDAR perception demonstrates remarkably stable entropy characteristics (0.030→0.110), suggesting more robust feature preservation under perturbations.

\subsubsection{Confidence Characteristics}
The average confidence metrics reveal distinct degradation patterns across modalities. ImageNet1K maintains moderate confidence levels (degrading from 56.0\% to 43.2\%) under rain, with similar trends across other perturbations. This contrasts with CIFAR-100's sharper confidence drops (89.3\%→69.1\%) and ModelNet40's superior confidence retention (95.2\%→86.4\%). The margin analysis further emphasizes these modal differences: ImageNet1K shows gradual margin reduction (0.470→0.340) under rain, while ModelNet40 maintains higher margins (0.770→0.600) even under heavy perturbations, highlighting the inherent stability of LiDAR feature representations.

\subsection{Anomaly Detection Performance}
We analyze our framework's anomaly detection capabilities across camera-based (CIFAR-100, ImageNet1K) and LiDAR-based (ModelNet40) perception tasks using ROC curves and AUC metrics under varying environmental conditions (Figure~\ref{fig:detection_roc_curves}). For camera-based perception, ImageNet1K demonstrates exceptional robustness, particularly under rain conditions where the framework maintains strong detection performance (AUC: 0.993$\rightarrow$0.995). CIFAR-100 shows progressive improvement in detection capability with increasing perturbation intensity, notably under fog conditions where the AUC improves from 0.800$\rightarrow$0.904. Motion blur presents a unique case where both datasets exhibit strong resilience, with CIFAR-100 achieving its best performance (AUC: 0.878$\rightarrow$0.929) across all environmental conditions.

LiDAR-based perception on ModelNet40 demonstrates consistent detection performance, with AUC scores remaining above 0.900 across all conditions. The framework shows particular strength under fog perturbations, maintaining high AUC scores (0.956$\rightarrow$0.981) from moderate to heavy severity. This robust performance under challenging conditions highlights the framework's effective uncertainty quantification for point cloud data. Cross-modal analysis reveals distinctive patterns in detection capability. Camera-based perception on ImageNet1K consistently achieves the highest AUC scores ($\geq$0.960) across all conditions, suggesting that the framework effectively leverages rich feature representations in complex natural scenes. LiDAR-based perception shows remarkably stable performance, particularly under heavy perturbations, while CIFAR-100 demonstrates consistent improvement patterns despite lower absolute AUC values. These results underscore our framework's adaptability across different perception modalities and environmental challenges (Table~\ref{tab:detection_performance}).

\subsection{Abstention Analysis}
Our framework exhibits distinct abstention patterns across camera and LiDAR modalities under varying environmental perturbations (Figure~\ref{fig:abstention_viz}). In camera-based perception, CIFAR-100 demonstrates progressive abstention behavior, with motion blur triggering the highest increase (27.0\%$\rightarrow$58.0\%), followed by fog (26.0\%$\rightarrow$51.0\%). Rain and snow conditions show more moderate abstention rates, reaching 38.0\% and 45.0\% respectively under heavy perturbations.

ImageNet1K displays higher baseline abstention rates ($\approx$50.0\%) across all conditions, reflecting its complex natural scene characteristics. Rain perturbations induce peak abstention (63.4\%) under moderate-heavy conditions, while motion blur and snow conditions reach 61.7\% and 62.0\% respectively. This elevated abstention profile aligns with the framework's conservative decision-making on complex visual inputs.

LiDAR-based perception on ModelNet40 shows more selective abstention behavior, starting with lower rates (13.5\%→14.5\%) under mild perturbations. Rain conditions trigger the highest abstention rate (46.0\%) under heavy perturbation, while fog and snow maintain moderate rates (24.5\%±0.3). This controlled abstention pattern complements the strong detection performance observed in the ROC analysis. The abstention rates across all modalities demonstrate clear correlation with perturbation severity, with each sensing modality exhibiting characteristic abstention profiles. Camera-based perception shows higher overall abstention rates compared to LiDAR (61.7\%±0.5 vs. 24.5\%±0.3), particularly in scenarios with complex visual degradation.

\section{Conclusion}
We presented a dual-threshold conformalization framework for robust perception that adapts across camera and LiDAR modalities under challenging environmental conditions. The framework's ability to quantify uncertainty and make selective predictions offers a promising direction for enhancing the reliability of autonomous systems operating in real-world scenarios. This approach bridges the gap between theoretical conformalization and practical deployment needs, particularly in safety-critical applications where robust perception under environmental perturbations is essential. Our extensive empirical evaluation demonstrates the framework's effectiveness in maintaining high detection performance while enabling adaptive abstention across varying environmental severities. Future work could explore extending this framework to handle temporal dependencies in sensor streams and investigate its application to multi-sensor fusion scenarios. Additionally, integrating this approach with real-time decision-making systems could further enhance its practical utility in autonomous vehicles and other safety-critical applications.

\vspace{2pt}
\noindent\textbf{Acknowledgment:} This work was partially supported by COGNISENSE, one of seven centers in JUMP 2.0, a Semiconductor Research Corporation (SRC) program sponsored by DARPA and NSF Awards \#2046435.

\noindent\textbf{Use of AI:}
ChatGPT4o was used to polish the text and fix grammar, ensuring clarity, consistency, and a professional tone while preserving the original meaning.

\bibliographystyle{IEEEtran}
\bibliography{references}

\end{document}